\documentclass[10pt,journal,compsoc]{IEEEtran}

\ifCLASSOPTIONcompsoc
  \usepackage[nocompress]{cite}
\else
  \usepackage{cite}
\fi
\ifCLASSINFOpdf
\else
\fi

\usepackage{amsmath,amsfonts}
\usepackage{algorithmic}
\usepackage{algorithm}
\usepackage{array}
\usepackage{url}
\usepackage{verbatim}
\usepackage{graphicx}
\usepackage{cite}
\usepackage{booktabs}
\usepackage{multirow}
\usepackage{xspace}

\newcommand{\eg}{{\emph{e.g.}},\xspace}

\usepackage{subfig}

\usepackage{bbding}

\usepackage{hyperref} 
\hypersetup{
    colorlinks=True,
    linkcolor=magenta,
    filecolor=magenta,      
    urlcolor=magenta,
    citecolor=cyan,
}

\newcommand{\INDSTATE}[1][1]{\STATE\hspace{#1\algorithmicindent}}
\usepackage{color}
\definecolor{GG}{RGB}{48,144,96}

\usepackage{subfig} 

\begin{document}
%
\title{Adaptive Outlier Optimization for Test-time Out-of-Distribution Detection}
%
%
%
%

\author{Puning~Yang, Jian~Liang, Jie~Cao, and~Ran~He\\

\IEEEcompsocitemizethanks{\IEEEcompsocthanksitem 
Puning Yang, Jian Liang, Jie Cao, Ran He are with the State Key Laboratory of Multimodal Artificial Intelligence Systems, CASIA, Center for Research on Intelligent Perception and Computing, CASIA, Center for Excellence in Brain Science and Intelligence Technology, CAS, and School of Artificial Intelligence, University of Chinese Academy of Sciences, Beijing 100190, China. E-mail: \{puning.yang, jie.cao\}@cripac.ia.ac.cn, liangjian92@gmail.com, rhe@nlpr.ia.ac.cn. (Corresponding author: Ran He.)}
\thanks{Preprint}}

%
%

\markboth{Journal of \LaTeX\ Class Files,~Vol.~14, No.~8, August~2015}%
{Shell \MakeLowercase{\textit{et al.}}: Bare Advanced Demo of IEEEtran.cls for IEEE Computer Society Journals}
%



\IEEEtitleabstractindextext{%
\begin{abstract}
Out-of-distribution (OOD) detection aims to detect test samples that do not fall into any training in-distribution (ID) classes. Prior efforts focus on regularizing models with ID data only, largely underperforming counterparts that utilize auxiliary outliers. However, data safety and privacy make it infeasible to collect task-specific outliers in advance for different scenarios. Besides, using task-irrelevant outliers leads to inferior OOD detection performance. To address the above issue, we present a new setup called test-time OOD detection, which allows the deployed model to utilize real OOD data from the unlabeled data stream during testing. We propose Adaptive Outlier Optimization (AUTO) which allows for continuous adaptation of the OOD detector. Specifically, AUTO consists of three key components: 1) an in-out-aware filter to selectively annotate test samples with pseudo-ID and pseudo-OOD and ingeniously trigger the updating process while encountering each pseudo-OOD sample; 2) a dynamic-updated memory to overcome the catastrophic forgetting led by frequent parameter updates; 3) a prediction-aligning objective to calibrate the rough OOD objective during testing. Extensive experiments show that AUTO significantly improves OOD detection performance over state-of-the-art methods. Besides, evaluations on complicated scenarios (\eg multi-OOD, time-series OOD) also conduct the superiority of AUTO.

\end{abstract}

}

\maketitle

\IEEEdisplaynontitleabstractindextext

%
\IEEEpeerreviewmaketitle

\ifCLASSOPTIONcompsoc
\IEEEraisesectionheading{\section{Introduction}\label{sec:introduction}}
\else
\section{Introduction}
\label{sec:introduction}
\fi

%
%
%
%
\IEEEPARstart{D}{eep} learning models have made tremendous progress in the past few years~\cite{lecun2015deep}.
The primitive to their success is that all classes that appear during inference are present in the training set.
However, such an assumption is criticized as it cannot always be satisfied in real-world applications where all the classes in the test phase would be available in the training phase.
Moreover, It is generally acknowledged that machine learning models often exhibit overconfident predictions on test samples that do not belong to any training classes~\cite{nguyen2015deep,bendale2016towards}.
This triggers a significant matter for robust, trustworthy, and safe AI applications, especially in 
autonomous driving~\cite{filos2020can}, fraud detection~\cite{phua2010comprehensive}, and medical diagnosis~\cite{roy2022does}.
To address the issue, recent efforts focus on out-of-distribution (OOD) detection~\cite{hendrycks2016baseline}.

\begin{table}[ht]
    \centering
    \caption{Comparisons of different detection paradigms.
    $\mathcal{D}^{aux}$ and $\mathcal{D}^{test}$ are datasets sampled from auxiliary data and test data, respectively.
    \textbf{Update}: Does model change at test time?
    \textbf{Online}: Does model continuously update?}
    \begin{tabular}{lccc}
    \toprule
       \multirow{2}{*}{Methods}  & \multicolumn{1}{c}{Training}& \multicolumn{2}{c}{Testing} \\
       &OOD&Update&Online\\
       \midrule
        Score Designing~\cite{hendrycks2016baseline} & -&\XSolidBrush&\XSolidBrush\\
        Outlier Exposure~\cite{hendrycks2018deep}&$\mathcal{D}^{aux}$&\XSolidBrush&\XSolidBrush\\
      WOODS~\cite{katzsamuels2022training}&$\mathcal{D}^{test}$&\XSolidBrush&\XSolidBrush\\
        Test Clustering~\cite{zhou2021step}&-&\Checkmark&\XSolidBrush\\
        Test-Time OOD Detection&-&\Checkmark&\Checkmark\\
        \bottomrule
    \end{tabular}
    
    \label{compare_methods}
\end{table}

The goal of OOD detection is to predict whether a test example is from a different distribution from the training data.
Existing solutions involve designing a scoring function that maps the input to the OOD score~\cite{hendrycks2016baseline,liang2018enhancing,lee2018simple,liu2020energy,hendrycks2022scaling} or modifying the training loss to mitigate the overconfident problem~\cite{wei2022mitigating,sehwag2020ssd,ming2023cider}.
Recent findings empirically suggest that exposing models to auxiliary outliers ($\mathcal{D}^{aux}$)~\cite{hendrycks2018deep,liu2020energy,ming2022posterior} significantly outperform the counterparts optimizing with ID data only, highlighting the potential of utilizing extra OOD data.
However, outliers collected offline may not closely match the true distribution of OOD data in deployments, failing to detect test OOD data.
Despite some efforts~\cite{katzsamuels2022training,zhou2021step,fan2022simple} that utilize test data for optimization, there remain some non-negligible limitations.
Firstly, previous test-time explorations require accessibility of test data, which is often impractical due to concerns related to data privacy and security.
Secondly, these explorations require utilizing mounts of test samples at one moment, which is difficult to achieve in online data streams.
Besides, all methods mentioned above are still evaluated in a simplistic offline test setting, where models remain fixed and static at test time (as shown in Table \ref{compare_methods}).
Such a restrictive setting hinders OOD detection in real-world environments, where wild data arrive sequentially and the OOD portion of the data is unknown and even ever-shifting.
For instance, in autonomous driving tasks, the deployed system struggles to perform well in an open environment where wild data arrive sequentially and gradually change with respect to geographic locations, time intervals, and other factors \cite{hoffman2014continuous}.
This motivates us to shift our perspective on OOD detection from the previous stationary to a dynamic setting.

In this paper, we pioneer a more practical setting called \emph{Test-Time OOD Detection (TTOD)}, where OOD detectors dynamically adapt to the current deployment scenario via continuous optimizations on unlabeled test data.
Specifically, the test samples in our setting arrive sequentially from either ID or OOD, inspiring a learner goal that incrementally updates the ID classifier and OOD detector based on predicted results of samples, and minimizes the risk of making incorrect predictions at each timestep.
Besides, the components in TTOD are more complex than those in the naive setting: (1) The ID and OOD mixture ratio are more flexible.
(2) More complex label shifts are considered, which include not only single-OOD scenarios but also multi-OOD and time-series OOD scenarios.
In contrast to the naive OOD detection setting, TTOD brings the benefits of practicality:
(1) Comprehensive: TTOD emulates various realistic scenarios, making it amenable to diverse real-world applications.
(2) Approximate: the online data stream provides a source of true distribution shift information, enabling TTOD to tailor the learning of an OOD detector for the deployed environment.

After formalizing TTOD, we further present a framework called Adaptive oUTlier Optimization (AUTO).
The framework leverages encountered test samples in real-time to perform targeted optimizations on the deployed model, aiming to perform better in making predictions when subsequent test samples arrive.
It comprises three key components: an in-out-aware filter, a dynamic-updated memory bank, and a prediction-aligning objective.
Firstly, to annotate unlabeled test samples as pseudo-ID or pseudo-OOD, we design an in-out-aware filter which is initialized with the statistics of model prediction confidence on ID data.
This filter predicts each test sample and selectively makes annotations, preparing for the optimization process.
Then, based on whether the annotation is pseudo-ID or pseudo-OOD, we directly use the corresponding objective to regularize models.
However, constant iterations lead to significant ID degradation, which is known as catastrophic forgetting.
Thus, we introduce a category-balanced ID memory bank, which contains one sample in each ID class and is updated with pseudo-ID data.
We design an OOD-triggered update strategy, which simultaneously updates model parameters with pseudo-OOD and the memory bank.
Last but not least, we observe that the previous paradigm makes an inappropriate objective on OOD data, which neglects the differences between OOD samples.
Therefore, we design a novel prediction-aligning objective, which slightly adjusts the OOD objective to fit the model’s intuition via aligning the prediction between the initial and current models.

To verify the efficacy of AUTO, we conduct extensive experiments on common CIFAR-10, CIFAR-100, and ImageNet benchmarks.
Except for naive OOD detection settings, we evaluate AUTO on the aforementioned challenging scenarios where test OOD data consists of multiple OOD data or time-series OOD data.
Natural language processing benchmarks are also included to validate the generality of our framework.
The results empirically demonstrate that our framework is capable of capturing the underlying OOD samples and maintaining expertise on ID tasks in learned latent spaces simultaneously for various OOD detection.
In summary, our work has the following contributions:
\begin{itemize}
        \item We firstly formalize the \emph{test-time out-of-distribution detection} setup.
        To the best of our knowledge, our work is the first to explore flexibly utilizing the test data stream for enhancing OOD detection.
        \item We present a new framework, Adaptive oUTlier Optimization, which adaptively mines distinct ID and OOD samples in the test stream while constantly optimizing with them.
        \item We design a novel prediction-aligning criterion, which calibrates the model's predictions for OOD data, resulting in better ID and OOD performance.
        \item More complex scenarios (\eg multi-OOD, time-series OOD) are considered within TTOD, providing a more comprehensive evaluation of OOD detection methods.
        Extensive experiments demonstrate the superiority of AUTO over other methods.
\end{itemize}


\section{Related Work}

\subsection{Out-of-Distribution Detection}
OOD Detection has been intensively studied in recent years~\cite{hendrycks2016baseline,liang2018enhancing,liu2020energy,huang2021mos,sun2022knn,du2023dream}.
Existing OOD detection methods typically train an offline supervised model on the in-distribution (ID) data, and then derive the OOD detector based on the learned classifier. 
In general, we can roughly categorize them into three categories according to their requirements for training data.

The initial category is the score design method, which doesn't necessitate specific training data. Its objective is to construct a scoring function that relates the input to the OOD score, signifying the degree to which the sample is considered out-of-distribution.
The OpenMax score~\cite{bendale2015towards} is the first method to detect OOD samples using the Extreme Value Theory.
Hendrycks \emph{et al.}~\cite{hendrycks2016baseline} present a baseline using the Maximum Softmax Probability (MSP) but may not be suitable for OOD detection \cite{morteza2022provable}.
Various scoring functions have been proposed to seek the properties that better distinguish OOD samples.
These new functions are calculated mainly from the output of the model.
Specifically, some of them are logit-based scores, such as  Energy score \cite{liu2020energy,wang2021canmulti,lin2021mood}, MaxLogit score \cite{hendrycks2022scaling}, DML score \cite{zhang2023decoupling}, and GEN score \cite{liu2023gen}.
Others are feature-based scores, such as Mahalanobis score \cite{lee2018simple}, GradNorm score \cite{huang2021importance}, and FeatureNorm score \cite{yu2023block}.
Except for general score designing literature, some works aim to enhance the aforementioned scores, such as ODIN score~\cite{liang2018enhancing} and React score~\cite{sun2021react}.

Except for designing scoring functions, the second category is modifying the logit space with novel loss functions, which only requires ID data during training~\cite{sehwag2020ssd,sun2022knn,morteza2022provable,wei2022mitigating,sun2022dice,ming2023cider,zhang2023decoupling,liu2023gen}.
Sun \emph{et al.}~\cite{sun2022knn} replace the classifier with a k-nearest neighbors predictor, thus eliminating the previous assumption about the distribution of feature space.
Ming \emph{et al.}~\cite{ming2023cider} propose CIDER, which jointly optimizes two losses to promote strong ID-OOD separability: a dispersion loss that promotes large angular distances among different class prototypes and a compactness loss that encourages samples to be close to their class prototypes.
Zhang \emph{et al.}~\cite{zhang2023decoupling} utilize the focal loss and the center loss, retraining networks to enhance the DML score.

Last but not least, outlier exposure methods, which require ID and auxiliary OOD data during training, can significantly enhance OOD detection performance.
The pioneering work, Outlier Exposure (OE) \cite{hendrycks2018deep}, optimizes the predictions of auxiliary OOD samples to a uniform distribution, inspiring a new line of work~\cite{liu2020energy,meinke2019towards,mohseni2020self,chen2021robustifying,ming2022posterior}.
Liu \emph{et al.}~\cite{liu2020energy} start from the energy perspective, optimizing OOD data to a higher-energy range.
Ming \emph{et al.}~\cite{ming2022posterior} propose a posterior sampling-based outlier mining framework, learning a more compact decision boundary than that of naive OE.
However, the main challenge of this paradigm lies in obtaining available auxiliary OOD data, which has inspired explorations into data augmentation~\cite{wang2023outofdistribution} and generation~\cite{du2022vos,tao2023nonparametric,du2023dream}.
For instance, Wang \emph{et al.}~\cite{wang2023outofdistribution} generate more outliers with data-augmentation methods, covering wider OOD situations than that of naive OE.
Except for auxiliary OOD data, some works focus on optimizations with test OOD data~\cite{katzsamuels2022training,zhou2021step,fan2022simple}.
For instance, Katzsamel \emph{et al.}~\cite{katzsamuels2022training} leverage the wild test data with an augmented-lagrangian constrained optimization, which makes the pre-trained model perform accurate OOD detection as well as ID classification.
Benefiting from auxiliary OOD data, this direction has demonstrated encouraging OOD detection performance compared to the aforementioned counterparts.



\subsection{Test-Time Optimization}
Except for modifying models during training, recent works explore the possibility of generalizing a pre-trained model to the target scenario during the inference phase \cite{liang2023comprehensive}.
Based on the training/test paradigm, existing works generally fall into three categories: source-free domain adaptation \cite{liang2020we}, test-time training \cite{sun2020test}, and test-time adaptation \cite{wang2020tent,iwasawa2021test}.

Source-free domain adaptation~\cite{liang2020we,nayak2021mining,li2021imbalanced,yang2021generalized} intends to transfer the source pre-trained model to the target domain during testing, utilizing all the test data in an offline manner.
Existing solutions leverage different techniques via pseudo-labeling~\cite{liang2018enhancing}, data generation~\cite{nayak2021mining}, memory bank~\cite{yang2021generalized}, clustering~\cite{li2021imbalanced,liu2022source}, and self-supervision~\cite{liang2021source,kundu2022concurrent}.
With sufficient adaptation to the target domain, these approaches always perform impressively.
However, it may be unavailable to access all the test data at once during testing, limiting the compatibility of the source-free setting.
Notably, some OOD detection works \cite{zhou2021step, fan2022simple} have explored similar source-free settings and achieved limited improvements.

Test-time training~\cite{sun2020test,gandelsman2022test,li2021test} introduces self-supervised auxiliary tasks during the training stage and optimizes them at test time to improve the performance of the source model.
To name a few, Sun \emph{et al.}~\cite{sun2020test} propose the pioneering work, which predicts the rotation angle~\cite{gidaris2018unsupervised} as the auxiliary task.
Gandelsman \emph{et al.}~\cite{gandelsman2022test} employ masked autoencoders~\cite{he2022masked} with vision transformer backbones to perform the self-supervised task.
Osowiechi \emph{et al.}~\cite{osowiechi2023tttflow} utilize unsupervised normalizing flows as an alternative auxiliary task.
Taking into account the inclusion of extra auxiliary tasks during training, these methods necessitate retraining the models, which ultimately renders them impractical for deployment in source-restricted scenarios.

Test-time adaptation endeavors to harness online, unlabeled test data streams to dynamically enhance the performance of the pre-trained model in real-time, which has garnered widespread attention in recent years~\cite{wang2020tent,iwasawa2021test,liu2021ttt++,jang2022test,chi2021test,kundu2020universal,royer2015classifier,niu2023towards,niu2022efficient,song2023ecotta}.
The early studies focus on calibrating the statistics of batch normalization layers~\cite{ioffe2015batch,nado2020evaluating}.
Afterwards, the following solutions pay attention to entropy minimization~\cite{wang2020tent}, pseudo-labeling updating~\cite{goyal2022test,jang2022test}, and output alignment~\cite{wang2022continual,dobler2023robust}.
Besides, some researchers attempt to achieve effective test-time adaption via parameter-free \cite{boudiaf2022parameter} or parameter-efficient \cite{gao2022visual} methods.

In short, prior test-time explorations largely focus on OOD generalization and overlook the OOD detection problem.
In contrast, our TTOD setup first fills the gap in OOD detection and test-time optimization.
It takes a more comprehensive consideration of evolving distribution shifts and online data streams and launches AUTO to tackle them by adaptively outlier annotation and optimization.

\subsection{Catastrophic Forgetting}
Forgetting refers to the loss or deterioration of previously acquired information or knowledge, which can be classified as harmful forgetting or beneficial forgetting.
In this paper, we focus on the harmful part, which has been observed not only in continual learning~\cite{parisi2019continual} but also in various other research areas, \eg domain adaptation~\cite{bobu2018adapting}, meta-learning~\cite{gidaris2018dynamic,finn2019online}, federated learning~\cite{karimireddy2020scaffold}, etc.
Considering that our research is related to test-time adaptation and continual learning, we introduce solutions about these areas, which can be divided into:
(1) Freezing the original parameters and introducing new learnable parameters to adapt the model to test-time data, such as VDP~\cite{gan2023decorate} and EcoTTA~\cite{song2023ecotta}. 
(2) Constraining the updates on crucial parameters to avoid introducing new parameters.
For instance, Tent~\cite{wang2020tent} only updates the batch normalization layers.
CoTTA~\cite{wang2022continual} randomly selects layers to update at each iteration.
EATA~\cite{niu2022efficient} updates parameters while calculating their importance to retain parameters crucial for the source domain.
Inspired by the aforementioned explorations, we intuitively overcome the forgetting issue with a memory bank, which is a common practice.
Furthermore, we design a prediction-aligning objective that provides new insights for addressing OOD detection.
This objective aligns the prediction between the initial and current models, improving ID performance effectively.

\subsection{Parameter-Efficient Fine-Tuning}
As the parameter number grows exponentially to billions \cite{brown2020language} or even trillions \cite{fedus2022switch}, it becomes very inefficient to save the fully fine-tuned parameters \cite{he2021towards} for each downstream task. Many recent research works propose a parameter-efficient \cite{houlsby2019parameter,zaken2022bitfit,he2021towards} way to solve this problem by tuning only a small part of the original parameters.
Parameter-efficient fine-tuning (PEFT) is initially proposed in natural language processing tasks~\cite{houlsby2019parameter} and later applied to computer vision tasks, which can be broadly grouped into addition-based approaches~\cite{he2021towards,houlsby2019parameter,jia2022visual,cai2019once,tu2023visual,sung2022lst,wu2021autoformer} and reparameterization approaches~\cite{caelles2017one,guo2021parameter,hu2021lora,lian2022scaling,yosinski2014transferable,zaken2022bitfit,zhao2020masking}.
PEFT is also a common practice in mounts of applications but has not been explored in OOD detection.
In this paper, we design an OOD-triggered strategy, which largely decreases iterations at test time.
Besides, our work is the first to discuss the impact of optimizing various components of the model parameters on the OOD detection performance.

\section{Test-Time OOD Detection}
In this section, we introduce the background of the OOD detection task (Section \ref{preli}) and provide a clear formulation of the test-time OOD detection setup (Section~\ref{setup_ttod}).

\subsection{Preliminaries: Naive OOD Detection}
\label{preli}
OOD detection is often formulated as a binary classification problem to distinguish between ID and OOD data.
We start from a multi-class classification task with input space $\mathcal{X} \subseteq \mathbb{R}^{d} $ and ID label space $\mathcal{Y}_{in}=\left \{ 1,...,C \right \}$, where $d$ and $C$ represent input dimensions and number of ID classes, respectively.
The supervised methods aim to learn the joint data distribution $\mathcal{P}_{\mathcal{X}\mathcal{Y}_{in}}$ from the labeled training set $\mathcal{D}_{tr}^{in} = \{x^{tr}_{i},y^{tr}_{i}\}_{i=1}^{N}$.

In the naive OOD detection setting, we are given a pre-trained model $f_{\theta_{0}}$ trained on $\mathcal{D}_{tr}^{in}$ will then be deployed in open-world scenarios containing OOD samples from unknown classes $y \notin \mathcal{Y}_{in}$.
Let $\mathcal{P}^{in}$ and $\mathcal{P}^{out}_t$ denote the marginal ID and OOD distribution on $\mathcal{X}$, respectively.
The open-world distribution $\mathcal{P}^{open}_t$ can be denoted with the Huber contamination model \cite{huber1992robust}:
\begin{equation}
    \mathcal{P}^{open}_t = \kappa_t\mathcal{P}^{in} + (1-\kappa_t)\mathcal{P}^{out}_t,
    \label{open_equ}
\end{equation}
where $\kappa_{t} \in [0,1]$ is a fixed value that presents the ID and OOD mixture ratio at $t$ time step.
The test set $\mathcal{D}_{te} = \{x^{te}_{i},y^{te}_{i}\}_{i=1}^{N}$ is sampled from $\mathcal{P}^{open}$, and it is noteworthy that all test samples $x^{te}_{i}$ can be accessed at once at any time step during the testing phase.
The goal of OOD detection is to decide if a test sample $x^{te} \in \mathcal{X}$ is from $\mathcal{P}^{in}$ or $\mathcal{P}^{out}$.
The decision process can be described via a thresholding comparison:
\begin{equation}
    D_{\beta}({x^{te}})=\begin{cases}
 {ID} & S({x^{te}}) \ge \beta \\
  {OOD}& S({x^{te}})<\beta
\end{cases},
\end{equation}
where $S(\cdot)$ is a score function and $\beta$ is the threshold.

\begin{figure}[h]
    \centering
    \includegraphics[width=\linewidth]{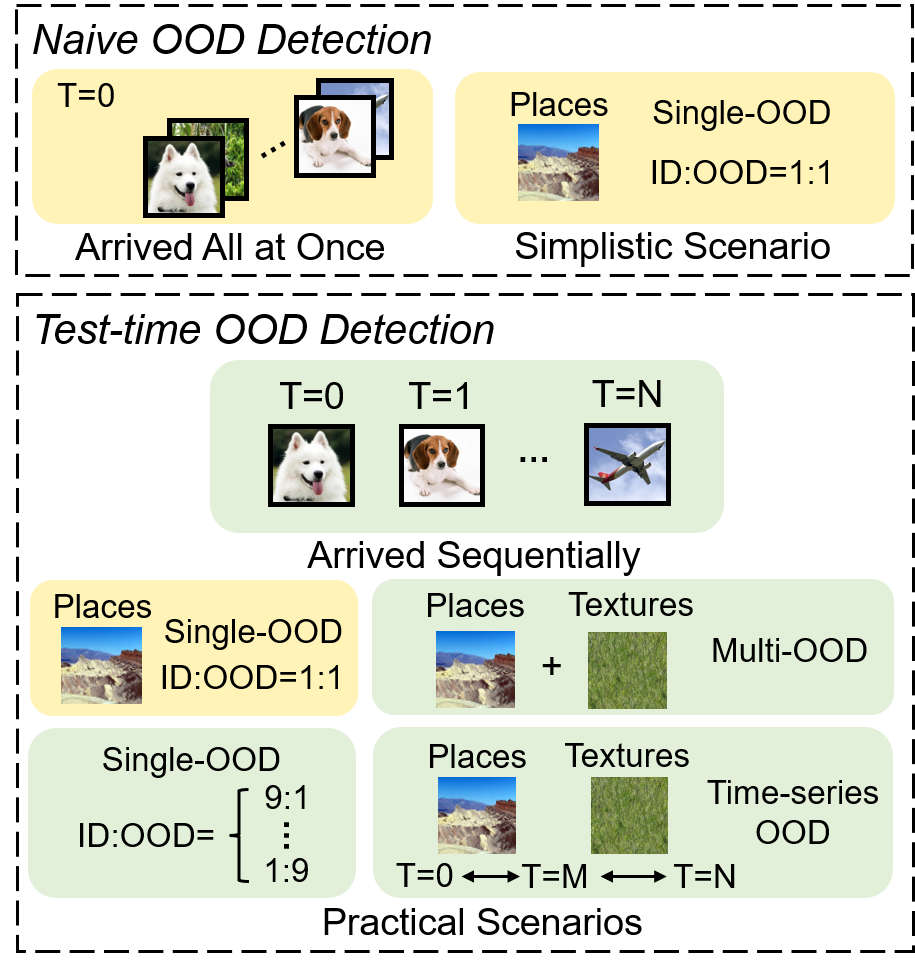}
    \caption{\textbf{Problem Formulation.} In contrast to naive OOD detection, we consider more practical and challenging scenarios, resulting in more comprehensive evaluations.}
    \label{fig:ttod_setting}
\end{figure}

\subsection{New Setup: Test-Time OOD Detection}
\label{setup_ttod}
As mentioned in the automation driving scenario, the naive OOD detection setting remains some impractical details. In the new test-time OOD detection, we make the following changes to the old setup, which emulate the more practical and challenging scenarios (as shown in Figure~\ref{fig:ttod_setting}):

1) \textbf{Accessing test data in an online manner.} Unlike the old offline setup, our new setup requires test data from $\mathcal{P}^{open}_t$ to arrive sequentially from either ID or OOD. Let $x^{te}_0, x^{te}_1, \cdots  ,x^{te}_N$ denote a stream of online unlabeled test samples, where $x^{te}_t$ is the test sample accessed at $t$ time step.

2) \textbf{Continous evolving data distribution.} Unlike the fixed $\kappa_{t}$, our new setting requires diverse (even dynamic) $\kappa_{t}$ to represent non-stationary environments.
Regarding the choice of $\kappa_{t}$, existing literature~\cite{katzsamuels2022training} argues that when the model is deployed in an open environment, the encountered OOD data will be much more than ID data.
Thus, $\kappa_{t}$ should be smaller.
Meanwhile, another~\cite{fan2022simple} believes that in practical applications, the model will likely be deployed in a more familiar environment where ID data will be much more than OOD data.
Thus, $\kappa_{t}$ should be larger.
Based on the above perspectives, we will comprehensively evaluate the performance of AUTO across a broad range of $\kappa_{t}$.

3) \textbf{More intricate OOD components.} 
Except for diverse $\kappa_{t}$, we consider more complex OOD components.
Let $\mathcal{P}^{OOD1}$ and $\mathcal{P}^{OOD1}$ denote different OOD distributions.
Unlike the old single-OOD scenario $\mathcal{P}^{out}_t \subset \mathcal{P}^{OOD1}$, our new setting considers multi-OOD $\mathcal{P}^{outM}_t$ and time-series OOD scenarios $\mathcal{P}^{outT}_t$:
\begin{equation}
    \mathcal{P}^{outM}_t \subset (\mathcal{P}^{OOD1} \cup \mathcal{P}^{OOD2})
\end{equation}
\begin{equation}
   \mathcal{P}^{outT}_t \subset \begin{cases}
   {\mathcal{P}^{OOD1}}& t \in [0,m] \\
 {\mathcal{P}^{OOD2}} & t \in (m,N]
\end{cases},
\end{equation}
where $m$ is a middle-time step during testing.

\begin{figure*}[ht]
    \centering
    \includegraphics[width=0.95\textwidth]{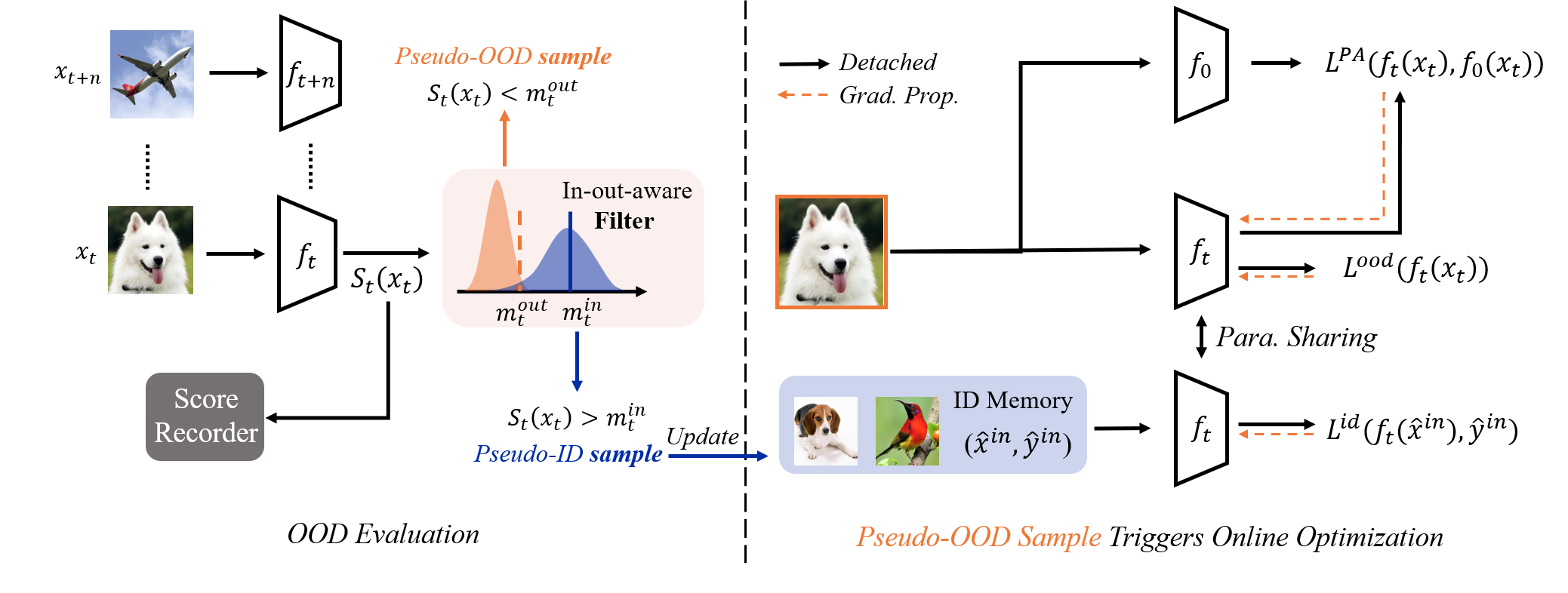}
    \caption{\textbf{Illustration of the Adaptive oUTlier Optimization (AUTO) framework.} The key components include an in-out-aware filter, a dynamic ID memory bank, and a prediction-aligning objective.
    Different color means different operations at test time: Each sample is given the MSP score and judged by the filter.
    Then, according to the judgment, the sample will activate different operations.
    For instance, if it is recognized as a pseudo-ID sample, blue lines are activated:
    this sample will be utilized to replace the sample with the same label in the ID memory bank.}
    \label{fig: framework}
\end{figure*}

\section{Method: Adaptive Outlier Optimization}
\label{auto_method}
In this section, we first introduce the proposed Adaptive oUTlier Optimization (AUTO) framework. As illustrated in Figure~\ref{fig: framework}, AUTO comprises three key components: an in-out-aware filter to tackle the selection of training samples (Section \ref{4.1}), a dynamic-updated ID memory bank, and a prediction-aligning objective to tackle the forgetting issue (Section \ref{4.2}), respectively.
Then, we elaborate on the parameter updating strategy for efficient model optimization (Section \ref{4.3}).
Last but not least, the full framework is provided with the above components, which systematically work as a whole and reciprocate each other (Section \ref{4.4}).

\subsection{Adaptive In-Out-Aware Filter}
\label{4.1}

Considering that ID and OOD samples have different optimization objectives, our intuition to utilize online test data is to annotate samples with pseudo-ID and pseudo-OOD.
Extensive prior works \cite{hendrycks2016baseline,liu2020energy,zhou2021step} have indicated that OOD data and ID data exhibit distinct distributions in feature space. 
Therefore, we design an in-out-aware filter, which is initialized with the statistical information of ID data in the softmax space.
For each incoming test sample, the filter can estimate the distance between this sample and the ID space, enabling rough annotations to be made.
Specifically, 
given ID examples ${x}_{i}^{in} \sim \mathcal{P}^{in}~,i \in [1,N]$, we compute the MSP \cite{hendrycks2016baseline} score $S_0({x}_{i}^{in})$ of each sample and then estimate the mean $\mu^{in}$ and standard deviation $\sigma^{in}$ of the ID data:
\begin{equation}
    \mu^{in}=\frac{\sum_{i=1}^{N}S_0({x}_{i}^{in})}{N}, \sigma^{in}=\sqrt{\frac{\sum_{i=1}^{N}(S_0({x}_{i}^{in})-\mu^{in})^2}{N}}. \\
\end{equation}
Then, the outlier-aware and inner-aware margins are initialized as follows:
\begin{equation}
m^{in}_0=\mu^{in} + k^{in} \times \sigma^{in},~~~m^{out}_0=\mu^{in} - k^{out} \times \sigma^{in},
\end{equation}
where $k^{in}$ and $k^{out}$ are hyper-parameters. 
\textbf{We can regard a sample with a score higher than $m^{in}$ as a pseudo-ID sample $(\hat{x}_t^{in},\hat{y}_t^{in})$, and a sample with a score lower than $m^{out}$ as a pseudo-OOD sample $\hat{x}_t^{out}$.}

\begin{figure}[h]
    \centering
    \includegraphics[width=\linewidth]{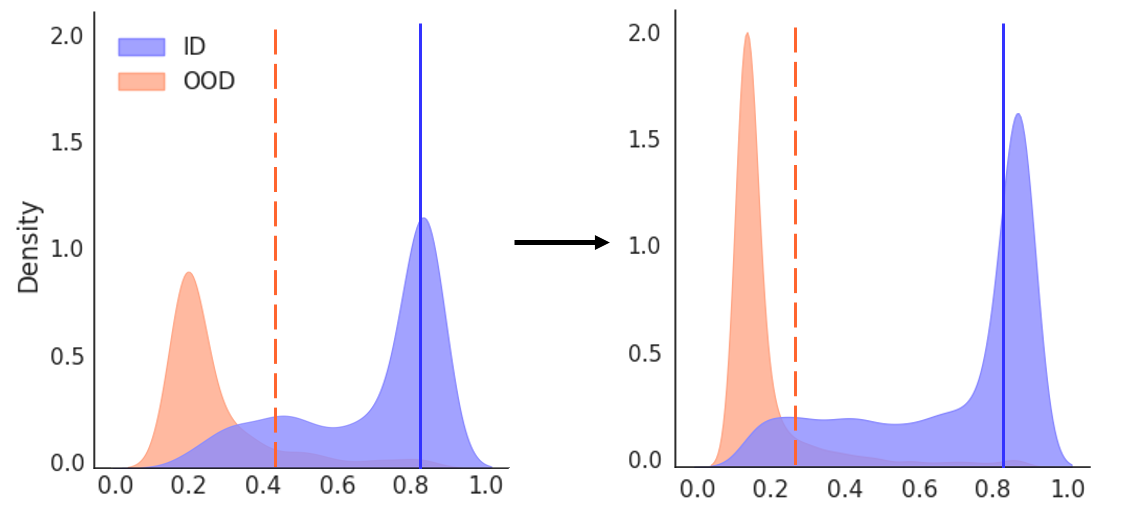}
    \caption{The distribution of MSP statistics is changing during testing, thus we update the OOD-aware margin and keep the ID-aware margin fixed.}
    \label{fig:filter}
\end{figure}

During the continuous updating process, the distribution of test data in the feature space undergoes constant changes. A common phenomenon~\cite{hendrycks2016baseline,liu2020energy,hendrycks2022scaling} is that the MSP scores of all samples are decreasing as we update models with outliers (as shown in Figure \ref{fig:filter}). 
Consequently, we have designed targeted update strategies for $m^{in}$ and $m^{out}$.
On the one hand, we keep $m^{in}$ fixed during training, which ensures that the labeling for pseudo-ID samples is correct.
On the other hand, we update $m^{out}$ with a greedy strategy.
We record the mean of historical OOD score values of the pseudo-OOD samples.
Then, we use the mean value to update $m^{out}$ as follows:
Assuming that we have recorded the mean score of $M$ pseudo-OOD samples when the t-th sample inputs:
\begin{equation}
    m^{out}_{t+1} =\begin{cases}
\frac{M \cdot m^{out}_{t} + S_{t}({x}_t)}{M+1}  & \text{ if } \ S_{t}({x}_t)<m^{out}_t ,\\
m^{out}_{t}  & \text{else.}
\end{cases}
    \label{outlier_margin}
\end{equation}

With the above annotations, the loss function for each pseudo-OOD sample $\mathcal{L}^{ood}_t$ can be defined through the cross-entropy between the prediction and target uniform vector:
\begin{equation}
 \mathcal{L}^{ood}_t=-\sum_{i=1}^{C} \frac{1}{C}\mathrm {log} \left (   \frac{\mathrm{exp}(f_{\theta_t}^{(i)}(\hat{x}_t^{out}))}{\sum_{j=1}^{C}\mathrm{exp}(f_{\theta_t}^{(j)}(\hat{x}_t^{out}))}\right ) ,
 \label{ood_loss}
\end{equation}
and the loss function for each pseudo-ID sample $\mathcal{L}^{id}_t$ can be defined with the cross-entropy loss:
\begin{equation}
 \mathcal{L}^{id}_t=-\mathrm {log} \left (   \frac{\mathrm{exp}(f_{\theta_t}^{(\hat{y}_t^{in})}(\hat{x}_t^{in}))}{\sum_{i=1}^{C}\mathrm{exp}(f_{\theta_t}^{(i)}(\hat{x}_t^{in}))}\right ).
 \label{id_loss}
\end{equation}
Leveraging these two loss functions defined above, we proceed to discuss how we can optimize the loss in the test data stream.




\subsection{Anti-Forgetting Components}

\begin{figure}[t]
    \flushleft
    \setlength{\abovecaptionskip}{10pt}
    \setlength{\belowcaptionskip}{-5pt}
    \subfloat[]{
        \label{fig: memory bank}
        \includegraphics[width=0.49\linewidth]{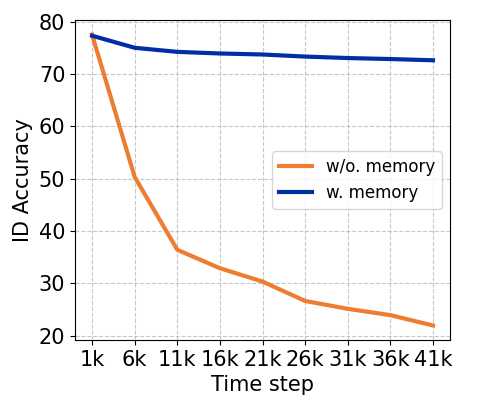}}
    \subfloat[]{
        \label{fig: predicting aligning}
        \includegraphics[width=0.49\linewidth]{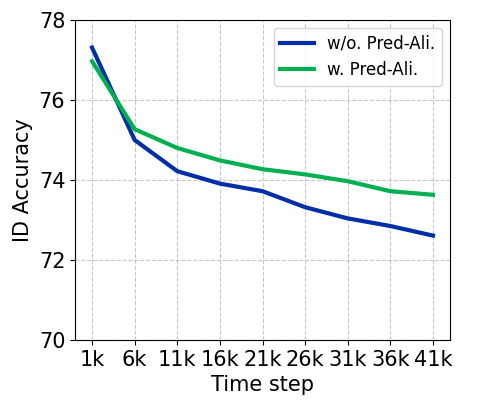}}
    \caption{\textbf{(a):} Models incur catastrophic forgetting due to constant updating, we mitigate the ID degradation with an ID memory.
    \textbf{(b):} We calibrate the objective of model and enhance ID and OOD performance further.}
    \label{fig:anti}
\end{figure}

To optimize the losses, one may intuitively think directly updates the OOD detector once it encounters $\hat{x}_t^{in}$ or $\hat{x}_t^{out}$.
However, as shown in Figure \ref{fig: memory bank}, we notice that such a simplistic optimization strategy significantly underperforms on the ID task.
Considering the constant parameters updating, we realize that the model encounters the catastrophic forgetting issue, a common phenomenon in online learning.
To address this issue, we upgraded the original alternating update strategy to a simultaneous update strategy.
Specifically, we introduce a dynamic memory bank and a prediction-aligning objective to mitigate ID degradation.
\label{4.2}

\textbf{Dynamic ID memory bank.}
We introduce a dynamic memory bank $\mathcal{M}^{id}$ into the ID classification loss formulated in Eq. \ref{id_loss}.
The memory bank stores one sample per category and is initialized with samples randomly selected from training data.
We update the samples in the memory bank with the test-time ID data in the same category.
Concretely, given a test-time sample $\hat{x}^{in}_t$ whose score is higher than the inner-aware margin $m^{in}$ and its pseudo label $\hat{y}^{in}_t$, we utilize it to update the memory bank as follows:
\begin{equation}
    \hat{x}^{in}_t \to {x}_{\mathcal{M}},\quad \text{if } \ \hat{y}^{in}_t= {y}_{\mathcal{M}}.
\end{equation}

Empirically, we notice that the training with only $\mathcal{M}_{id}$ does not help improve OOD detection.
Therefore, we design an OOD-triggered strategy that modifies the model only when encountering a pseudo-OOD sample $\hat{{x}}^{out}_t$, reducing iterations significantly.
We update the model with $\mathcal{L}^{ood}_t$ and $\mathcal{L}^{id}_t$ simultaneously.
As shown in Figure \ref{fig: memory bank}, our new optimization strategy largely mitigates the ID degradation.

Except for considerations focused on ID performance, we also notice the additional computational and time burden brought by the ID memory.
Thus, we introduce a part-activate strategy while ID memory is large.
For instance, we still update an ID memory that contains 1000 samples when evaluating AUTO on the ImageNet-1k dataset, but we only randomly activate 100 of them when we update models.

\begin{figure}[h]
    \centering
    \includegraphics[width=\linewidth]{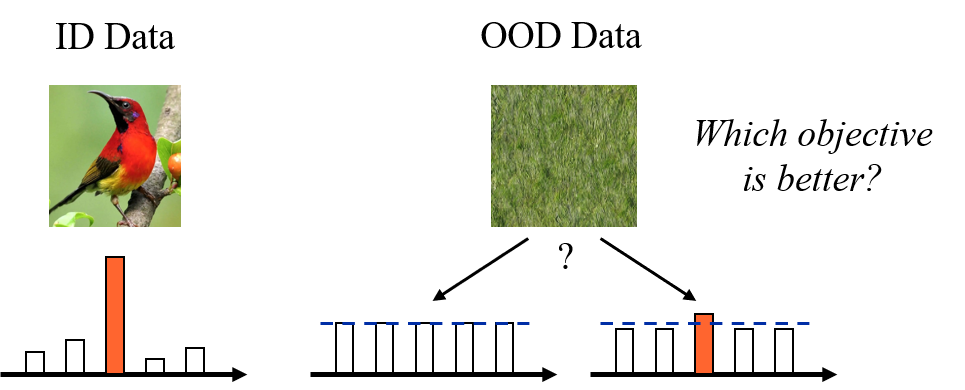}
    \caption{Calibration on the OOD objective. Based on a uniform vector, we propose to consider background information, leading to a more optimal objective that aligns with the model's intuition.}
    \label{fig:pa_ill}
\end{figure}

\textbf{Prediction-Aligning Objective.}
Except for reminding the OOD detector what it has learned before.
We notice that roughly aligning the predictions of all OOD samples with a uniform vector is inappropriate.
Considering the potential similarity between ID images and OOD images (as shown in Figure \ref{fig:pa_ill}), we believe the optimization objective should have slight adjustments based on the model's intuition.
Furthermore, we find that at the beginning of test-time optimization, $m^{out}_0$ may misclassify some ID samples as OOD.
Such misclassifications subsequently confuse the model during optimization.
To address this problem, we propose aligning the predictions of pseudo-OOD samples between the original model and the updated model.
Specifically, at the beginning of the testing stage, we make a duplicate of the model $f_{\theta_0}$ and freeze its parameters.
The prediction of the duplicated model is denoted as ${y}_0$.
Intuitively, if the results predicted by the model remain consistent with ${y}_0$, the performance on the source task will not degrade.
Let $p_t^{{y}}$ denote the softmax probability that the t-th sample belongs to the class ${y}$.
Our objective is:
\begin{equation}
    \mathcal{L}^{PA}_t=\begin{cases}
0,  & \text{ if } \ y_0=y_t \\
p_t^{y_t} - p_t^{y_0} + \phi, & \text{ if } \ y_0 \neq y_t
\end{cases},
\end{equation}
which enforces $p_t^{y_t}$ close to $p_t^{y_0}$ with a margin $\phi$. That means the prediction of $f_{\theta_t}$ is supposed to be higher than that of $f_{\theta_0}$ at least by $\phi$. 

\subsection{Parameter-Efficient Update}
\label{4.3}
Different from training-time methods, some real-world applications have requirements on the per-sample process time.
Thus, we attempt to accelerate the optimization of each test sample.
Let $\theta$ denote all the parameters of the model, updating $\theta$ is a natural choice, but it is sub-optimal for test-time OOD detection.
While part-parameter optimization is common in many tasks, there is still a lack of research on identifying which part should be updated to improve OOD detection performance efficiently.
Following the partial updating principle~\cite{wang2020tent}, we explore the influence of optimizing different combinations of parameters, \eg, the last feature block $\theta_{last}$, all batch normalization layers $\theta_{bn}$, and the classifier $\theta_{fc}$.
Table~\ref{tab:opti} displays the results that optimize the above combinations.
We finally optimize $\theta_{last}$ while keeping the remaining parameters fixed during testing.
Besides, OOD-triggered optimization is also an efficient strategy that has been mentioned in Section \ref{4.2}.

\subsection{Overall Objective}
\label{4.4}
Finally, the overall objective of AUTO is shown as:
\begin{equation}\mathcal{L}^{total}_t=\mathcal{L}^{id}_t+\lambda^{out}\mathcal{L}^{ood}_t+\lambda^{PA}\mathcal{L}^{PA}_t,
\label{totalloss}
\end{equation}
where $\lambda^{out}$ and $\lambda^{PA}$ are hyper-parameters.
To adequately leverage the information in the outlier and the memory bank,
we repeat the optimization process on each outlier iteratively for a given number of iterations, $T$.
While seemingly separated from each other, the three components of AUTO are working collaboratively.
First, the in-out-aware filter selects high-quality ID and OOD samples from the unlabeled test data, which facilitates the positive update of models.
Second, the anti-forgetting components help the model enlarge the margin between ID and OOD data, which pays back to the filter and helps it select samples more accurately.
The entire training process converges when the three components perform satisfactorily.

\section{Experiment Result and Discussion}
In this section, we evaluate AUTO for test-time OOD detection on computer vision (CV) and natural language processing (NLP) benchmarks.
We compare AUTO with previous OOD detection methods, both OOD performance and ID performance (Section \ref{exp_res}).
Besides, we present extensive ablation experiments of different components to understand their contribution toward the performance (Section \ref{abla}).

\subsection{Experimental Setup}
\label{setup}

\noindent \textbf{Datasets.}
Following the common CV benchmarks in OOD detection literatures~\cite{liu2020energy,du2022vos}, we evaluate our method on CIFAR-10/100 \cite{krizhevsky2009learning} and ImageNet-1k \cite{deng2009imagenet}.

For CIFAR benchmarks, we consider six common OOD datasets: \textbf{SVHN} \cite{netzer2011reading}, \textbf{Textures} \cite{cimpoi2014describing}, \textbf{LSUN-Crop} \cite{yu2015lsun}, \textbf{LSUN-Resize} \cite{yu2015lsun}, \textbf{iSUN} \cite{xu2015turkergaze}, and \textbf{Places} \cite{zhou2017places}.
We utilize \textbf{ImageNet-1k}~\cite{deng2009imagenet} as the auxiliary outliers.

For the ImageNet benchmark, we use subsets of four datasets from \textbf{SUN} \cite{xiao2010sun}, \textbf{Textures} \cite{cimpoi2014describing}, \textbf{Place} \cite{zhou2017places}, and \textbf{iNaturalist} \cite{van2018inaturalist}.
We utilize \textbf{ImageNet-22k}~\cite{codreanu2017scale} as the auxiliary outliers.
To make test OOD data and auxiliary outliers disjoint, images in ImageNet-1k are removed.

For NLP benchmarks, we evaluate our method on 20 Newsgroups~\cite{pedregosa2011scikit} and TREC~\cite{li2002experimental}. 
For each ID dataset, we consider five common OOD datasets: \textbf{SNLI} \cite{bowman2015large}, \textbf{IMDB} \cite{maas2011learning}, \textbf{Multi30K} \cite{elliott2016multi30k}, \textbf{WMT16}~\cite{bojar2016findings}, and \textbf{Yelp} \cite{asghar2016yelp}.
We utilize \textbf{Gutenburg}~\cite{csaky2021gutenberg}, \textbf{WikiText-2}~\cite{merity2016pointer}, and \textbf{WikiText-103}~\cite{merity2016pointer} as the auxiliary outliers.

Images in CIFAR and ImageNet benchmarks are resized to $32 \times 32$ and $224 \times 224$, respectively.
To simulate the online teat data stream, the batch size is set to 1.
We provide more details in Table \ref{tab:id_data} and \ref{imagenet:ood_data}.

\begin{table}[h]
    \centering
    \caption{Details of ID datasets.}
    \setlength{\belowcaptionskip}{-10pt}
    \begin{tabular}{c|cccc}
    \toprule
         ID Data& CIFAR10/100 &ImageNet-1k & 20-NG &TREC\\
    \midrule
         Training&50,000&1,281,167&11,293&5,452\\
         Testing&10,000&50,000&7,528&500\\
    \bottomrule
    \end{tabular}
    
    \label{tab:id_data}
\end{table}

\begin{table}[h]
    \centering
    \caption{Details of OOD datasets.}
    \resizebox{\linewidth}{!}{
    \begin{tabular}{c|ccccc}
    \toprule
    CIFAR & SVHN & Textures& Places& iSUN& LSUN\\
    OOD &10,000&5640&10,000&10,000&10,000\\
    \midrule
        ImageNet & iNaturalist & Textures& Places50& SUN\\
        OOD &10,000&5640&10,000&10,000\\
    \midrule
    NLP&SNLI&IMDB&Multi30K&WMT16&Yelp\\
    OOD&9,824&25,000&29,000&22,191&50,000\\
    \bottomrule
    \end{tabular}
    }
    
    \label{imagenet:ood_data}
\end{table}

\noindent \textbf{Backbones.}
For CIFAR benchmarks, we train two backbones from scratch: ResNet-34 \cite{he2016identity} and the Wide ResNet \cite{zagoruyko2016wide} architecture with 40 layers and a widen factor of 2.
The models are trained for 100 epochs. 
The start learning rate is to 0.1 which decays by a factor of 10 at epochs 50 and 80.
Batch size is set to 128 for backbones used in CIFAR benchmarks.
For the ImageNet benchmark, we use a pre-trained ResNet-50 model \cite{he2016identity} from the PyTorch \cite{paszke2019pytorch} and a pre-trained Vision Transformer \cite{dosovitskiy2020image} from the Timm library \cite{rw2019timm}.
For NLP benchmarks, we train a 2-layer GRUs \cite{cho2014learning} model for 5 epochs at training time.

\noindent \textbf{Implementation details.}
For OE-based methods, models are fine-tuned for 10 epochs in CV tasks and for 2 epochs in NLP tasks.
For modifying during testing, we use stochastic gradient descent with the learning rate set to that of the last epoch during training, which is 0.001 in all our experiments.
We set weight decay and momentum to zero during test-time OOD detection, inspired by practice in \cite{liu2018rethinking,he2019rethinking}.
Hyper-parameters are adjusted as backbone changes and shown in Table \ref{tab:hyper}.

\begin{table}[h]
    \centering
    \caption{Hyper-parameter setting of different backbones.}
    \begin{tabular}{l|ccccc}
    \toprule
         Model& $\lambda^{out}$ & $\lambda^{PA}$ & $\phi$& $k^{in}$& $k^{out}$\\
    \midrule
         ResNet-34& 0.25&0.2&0.05&0&3\\
         WRN-40-2&0.25&0.1&0.05&0&3\\
         ResNet-50&0.25&0.1&0.005&0&3\\
         ViT-B-16&0.25&0.1&0.005&0&1.5\\
         \bottomrule
    \end{tabular}
    
    \label{tab:hyper}
\end{table}


\noindent \textbf{Baselines.}
The compared algorithms include: (1) Methods train model without auxiliary outliers: MSP \cite{hendrycks2016baseline}, ODIN \cite{liang2018enhancing}, Mahalanobis \cite{lee2018simple}, Energy \cite{liu2020energy}, GradNorm \cite{huang2021importance}, MaxLogit~\cite{hendrycks2022scaling}, ViM~\cite{wang2022vim}, LogitNorm \cite{wei2022mitigating}, ReAct \cite{sun2021react}, DICE \cite{sun2022dice}, KNN \cite{sun2022knn}, and DML+~\cite{zhang2023decoupling}.
(2) Methods train model with auxiliary outliers: OE~\cite{hendrycks2018deep}, Energy~\cite{liu2020energy}, VOS \cite{du2022vos}, POEM~\cite{ming2022posterior}, WOODS~\cite{katzsamuels2022training}, NPOS \cite{tao2023nonparametric}, and DOE~\cite{wang2023outofdistribution}.
The reported results in this paper are obtained by reproducing the source code provided by the aforementioned methods.

\noindent \textbf{Evaluation metrics.}
We evaluate our framework and baseline methods using the following metrics: 1) The false positive rate of OOD samples when the true positive rate of in-distribution samples is at 95\% (\textbf{FPR95}); 2) The area under the receiver operating characteristic curve (\textbf{AUROC}); and 3) The ID classification accuracy (\textbf{ID\_Acc}).

\subsection{Performance Analysis}
\label{exp_res}

\setlength{\tabcolsep}{2pt}
\begin{table*}[ht]
    \caption{Comparison with competitive OOD detection methods on CIFAR benchmarks. $\uparrow$ indicates larger values are better and vice versa.
    All values are percentages averaged over six OOD test datasets described in Section \ref{setup}.
    Bold numbers indicate superior results.
    $\mathcal{D}^{aux}$ indicates whether the detector is modified with an outlier dataset during training.}
    \resizebox{\textwidth}{!}{
    \centering
    
    \begin{tabular}{lcrrrrrrrrrrrr}
    \toprule
    \makebox[0.1\textwidth][l]{\multirow{3}{*}{\textbf{Methods}}} & \multirow{3}{*}{$\mathcal{D}^{aux}$}&
    \multicolumn{6}{c}{\textbf{CIFAR-10}} & \multicolumn{6}{c}{\textbf{CIFAR-100}}\\
    & &\multicolumn{3}{c}{\textbf{ResNet-34}}&\multicolumn{3}{c}{\textbf{WideResNet-40-2}}&\multicolumn{3}{c}{\textbf{ResNet-34}}&\multicolumn{3}{c}{\textbf{WideResNet-40-2}}\\
    \cmidrule(lr){3-5}\cmidrule(lr){6-8}\cmidrule(lr){9-11}\cmidrule(lr){12-14}
    & &\textbf{FPR95}$\downarrow$&\textbf{AUROC}$\uparrow$&\textbf{ID\_Acc}$\uparrow$&\textbf{FPR95}$\downarrow$&\textbf{AUROC}$\uparrow$&\textbf{ID\_Acc}$\uparrow$&\textbf{FPR95}$\downarrow$&\textbf{AUROC}$\uparrow$&\textbf{ID\_Acc}$\uparrow$&\textbf{FPR95}$\downarrow$&\textbf{AUROC}$\uparrow$&\textbf{ID\_Acc}$\uparrow$\\
    \midrule
    MSP \cite{hendrycks2016baseline}&\XSolidBrush&46.49&92.53&94.87 &52.00&90.57&94.53 &83.53&74.34&77.51 &79.15&76.44&75.84\\
    ODIN\cite{liang2018enhancing}   &\XSolidBrush&30.00&93.94&94.87 &34.32&91.38&94.53 &82.76&75.27&77.51 &69.75&81.29&75.84\\
    Mahalanobis \cite{lee2018simple}&\XSolidBrush&44.31&93.31&94.87 &25.61&95.19&94.53 &75.56&80.82&77.51 &71.14&79.71&75.84\\
    Energy \cite{liu2020energy}     &\XSolidBrush&28.77&94.07&94.87 &33.41&91.53&94.53 &82.65&75.33&77.51 &69.65&81.30&75.84\\
    GradNorm \cite{huang2021importance}                &\XSolidBrush&66.20&83.21&94.87 &71.83&61.91&94.53 &76.45&71.46&77.51& 85.28&56.14&75.84\\
    MaxLogit \cite{hendrycks2022scaling}               &\XSolidBrush&28.05&94.02&94.87 &35.01&91.06&94.53 &80.67&77.39&77.51& 73.29&80.25&75.84\\
    ViM \cite{wang2022vim}                             &\XSolidBrush&26.21&94.73&94.87 &23.02&94.98&94.53 &66.22&82.36&77.51& 66.24&82.51&75.84\\
    ReAct \cite{sun2021react}                          &\XSolidBrush&32.57&93.16&94.85 &58.67&82.85&93.41 &74.76&82.01&77.09 &92.01&64.53&64.77\\
    Logit Norm \cite{wei2022mitigating} 
    &\XSolidBrush&18.14&96.61&94.68 &21.03&95.86&94.42 &76.08&76.83&76.40 &54.90&87.60&\textbf{76.02}\\
    KNN \cite{sun2022knn}&\XSolidBrush&36.71&94.15&94.87&36.63&93.31&94.53&71.33&82.44&77.51&59.92&84.36&75.84\\
    DML+\cite{zhang2023decoupling}  &\XSolidBrush&23.66&95.34&94.87 &10.08&98.00   &94.53   &58.02&87.53&77.51&39.20&91.21&75.84\\
    \midrule
    OE~\cite{hendrycks2018deep}&\Checkmark&6.02&98.58&95.08&7.08&98.51&94.44&58.52&87.30&76.84&54.04&85.82&75.59\\
    Energy ~\cite{liu2020energy}&\Checkmark&\textbf{2.93}&98.71&\textbf{95.49}&\textbf{2.91}&\textbf{98.97}&\textbf{94.91}&53.02&90.05&77.19&44.43&90.47&75.75\\
    POEM~\cite{ming2022posterior}&\Checkmark&11.25&97.62&89.57 &7.17&98.37&90.62 &19.78&95.94&69.49&24.30&95.96&69.37\\
    WOODS~\cite{katzsamuels2022training}&\Checkmark&10.10&97.75&94.79&12.14&97.58&94.72&34.90&91.21&77.84&22.65&94.54&75.74\\
    DOE~\cite{wang2023outofdistribution}&\Checkmark&8.93&97.84&94.74&5.00&98.75&94.43&32.43&93.65&76.95&26.09&94.43&74.98\\
    \midrule
    AUTO&\XSolidBrush&6.22&\textbf{98.72}&94.92&9.45&97.94&94.33&\textbf{11.06}&\textbf{97.50}&\textbf{77.91} &\textbf{16.45}&\textbf{95.97}&74.77\\
    \bottomrule
    \end{tabular}
    }
    \setlength{\belowcaptionskip}{-10pt}
    
    \label{tab:cifar}
\end{table*}

\begin{figure}[h]
    \centering
    \includegraphics[width=\linewidth]{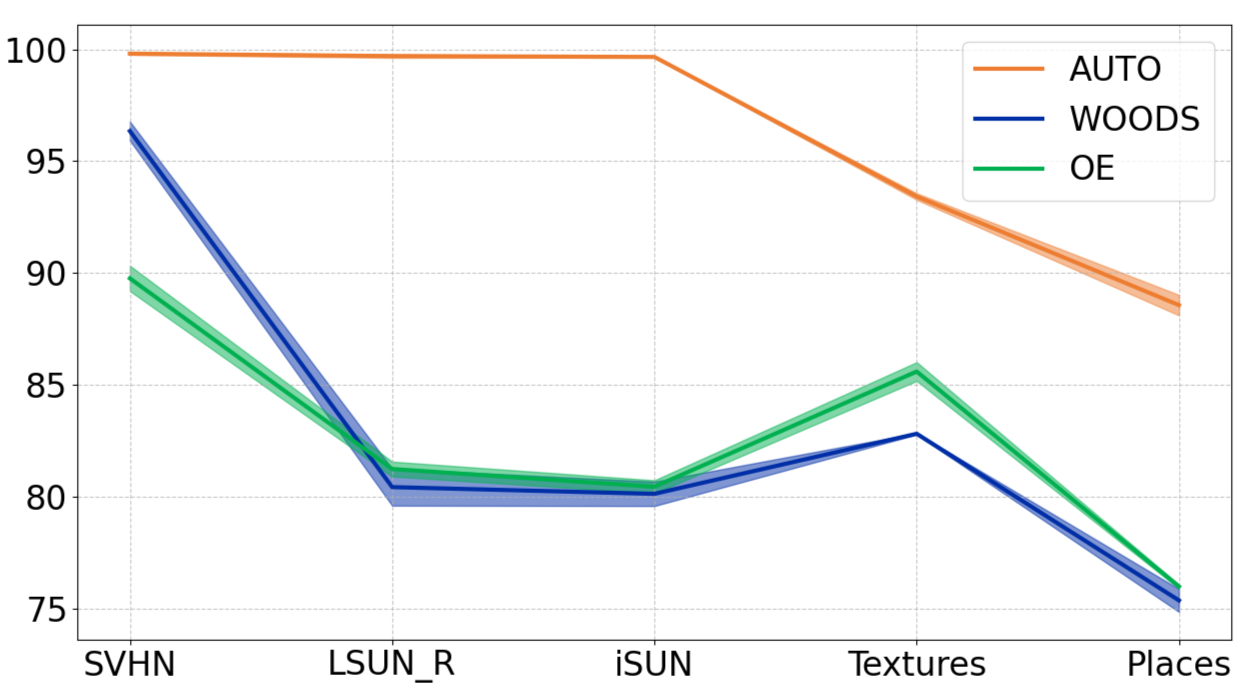}
    \caption{OOD performance on time-series OOD scenarios, AUROC is reported.
    ResNet-34 is trained on CIFAR-100.}
    \label{fig:tso}
\end{figure}

Extensive experiments and results are presented here.
Firstly, for fairness, the ratios of in-distribution (ID) to out-of-distribution (OOD) data in Tables \ref{tab:cifar} and \ref{imagenet} follow the same proportions as set in the naive OOD detection, mitigating the adverse impact of complex scenarios in test-time OOD detection on the performance of previous methods.
Then, major solutions are evaluated on new test scenarios (multi-OOD in Table \ref{tab:mix_data} and time-series OOD in Figure \ref{fig:tso}), which further present the superiority of AUTO.
Last but not least, evaluations on NLP benchmarks exhibit the generality and compatibility of AUTO.

\textbf{AUTO significantly outperforms counterparts that train models without auxiliary outliers.} 
Compared with methods that optimize models with only ID data, AUTO inherently exhibits superior OOD detection performance, which is attributed to modifications with real-OOD samples at test time.
Meanwhile, AUTO is training-free, maintaining the advantage of not requiring auxiliary outliers during training.
Part of previous explorations require a mount of ID data for retraining to overcome overconfidence.
In contrast, AUTO has no requirements, significantly alleviating the dependence of pretrained models on source data. 
These characteristics make AUTO more suitable for deployments.

\setlength{\tabcolsep}{3.0pt}
\begin{table*}[t]
    \centering
    \caption{Comparison with competitive OOD detection methods on the ImageNet benchmark.}
    \resizebox{\textwidth}{!}{
    \begin{tabular}{lcrrrrrrrrrrr}
    \toprule
    \multicolumn{1}{c}{\multirow{3}{*}{\textbf{Methods}}} & \multirow{3}{*}{$\mathcal{D}^{aux}$}& \multicolumn{8}{c}{\textbf{OOD Datasets}} & \multicolumn{2}{c}{\multirow{2}{*}{\textbf{Average}}}&\multirow{3}{*}{\textbf{ID\_Acc}$\uparrow$}\\ 
    \cmidrule(lr){3-10}
    \multicolumn{2}{c}{ } & \multicolumn{2}{c}{\textbf{SUN}}   & \multicolumn{2}{c}{\textbf{Textures}}  & \multicolumn{2}{c}{\textbf{iNaturalist}} & \multicolumn{2}{c}{\textbf{Places}}  \\ 
    \cmidrule(lr){3-4}\cmidrule(lr){5-6}\cmidrule(lr){7-8}\cmidrule(lr){9-10}\cmidrule(lr){11-12}
    \multicolumn{2}{c}{} &\multicolumn{1}{c}{FPR95$\downarrow$} &AUROC$\uparrow$  & FPR95$\downarrow$ &AUROC$\uparrow$  & FPR95$\downarrow$ &AUROC$\uparrow$  & FPR95$\downarrow$ & \multicolumn{1}{c}{AUROC$\uparrow$}  & \multicolumn{1}{c}{FPR95$\downarrow$ }&AUROC$\uparrow$ \\ 
    \midrule
    \multicolumn{13}{c}{\textbf{Backbone: ResNet-50}}\\
    MSP\cite{hendrycks2016baseline}&\XSolidBrush& 68.53 & 81.75 & 66.15 & 80.46 & 52.69 & 88.42 & 71.59 & 80.63 & 64.74 & 82.82 & \textbf{76.12} \\ 
    ODIN\cite{liang2018enhancing}&\XSolidBrush& 54.04 & 86.89 & 45.50 & 87.57 & 41.50 & 91.38 & 62.12 & 84.45 & 50.79 & 87.57 & \textbf{76.12} \\ 
    G-ODIN\cite{hsu2020generalized}&\XSolidBrush&60.83&85.60&77.85&73.27&61.91&85.40&63.70&83.81&66.07&82.02&\textbf{76.12}\\
    Mahalanobis\cite{lee2018simple}&\XSolidBrush&98.35&42.10&54.78&85.02&96.95&52.60&98.47&42.01&87.14&55.43&76.12\\
    Energy\cite{liu2020energy}&\XSolidBrush& 58.25 & 86.73 & 52.30 & 86.73 & 53.94 & 90.60 & 65.40 & 84.12 & 57.47 & 87.05 & \textbf{76.12} \\
    MaxLogit\cite{hendrycks2018deep}&\XSolidBrush& 60.42&86.44&66.05&84.03&50.82&91.15&54.95&86.39&58.06&87.00&\textbf{76.12}\\
    GradNorm\cite{huang2021importance}&\XSolidBrush&38.53&88.87&46.76&83.66&31.24&91.79&46.29&86.28&40.71&87.65&\textbf{76.12}\\
    ViM\cite{wang2022vim}&\XSolidBrush&91.87&72.65&12.40&97.52&67.95&88.40&91.09&71.47&65.83&82.51&\textbf{76.12}\\
    ReAct\cite{sun2021react}&\XSolidBrush& 23.69 & 94.44 & 46.33 & 90.30 & 19.71 & 96.37 & 33.30 & 91.96 & 30.76 & 93.27 & 74.82 \\ 
    DICE+ReAct\cite{sun2022dice}&\XSolidBrush&  26.49 & 93.83 & 29.36 & 92.65 & 20.07 & 96.11 & 38.35 & 90.61 & 28.57 & 93.30 & 67.01 \\ 
    KNN\cite{sun2022knn}&\XSolidBrush&70.50&80.46&11.26&97.41&60.30&86.09&78.81&74.66&55.22&84.66&\textbf{76.12}\\
    DML+\cite{zhang2023decoupling}&\XSolidBrush&30.73&93.98&36.35&89.02&13.66&97.48&39.82&91.22&30.14&92.93&\textbf{76.12}\\
    OE\cite{hendrycks2018deep}&\Checkmark& 80.10&76.55&66.38&82.04&78.31&75.23&70.41&81.78&73.80&78.90&75.51\\
    MixOE\cite{zhang2023mixture}&\Checkmark&74.62&79.81&58.00&85.83&80.51&74.30&84.33&69.20&74.36&77.28&74.62\\
    VOS\cite{du2022vos}&\Checkmark&98.72&38.50&70.20&83.62&94.83&57.69&87.75&65.65&87.87&61.36&74.43\\
    DOE\cite{wang2023outofdistribution}&\Checkmark&80.94&76.26&34.67&88.90&55.87&85.98&67.84&83.05&59.83&83.54&75.50\\
    AUTO&\XSolidBrush &\textbf{8.26} & \textbf{97.34} & \textbf{11.21} & \textbf{97.68} & \textbf{2.00} & \textbf{99.39} & \textbf{18.35} & \textbf{94.98} & \textbf{9.96} & \textbf{97.44} & 74.64 \\
    \midrule
    \multicolumn{13}{c}{\textbf{Backbone: ViT-Base-16}}\\
    MSP\cite{hendrycks2016baseline}&\XSolidBrush  & 73.80 & 79.49 & 63.07 & 81.50 & 39.40 & 92.41 & 74.09 & 79.56 & 62.59 & 83.24 & 78.01 \\ 
    ODIN\cite{liang2018enhancing}&\XSolidBrush  & 62.81 & 83.20 & 51.45 & 86.31 & 30.28 & 92.65 & 66.21 & 81.51 & 52.69 & 85.92 & 78.01 \\ 
    Mahalanobis\cite{lee2018simple}&\XSolidBrush&79.88&81.82&72.10&80.33&18.22&95.37&84.05&73.70&63.57&82.81&78.01\\
    Energy\cite{liu2020energy}&\XSolidBrush  & 69.29 & 84.52 & 51.97 & 88.30 & 37.84 & 94.46 & 72.03 & 82.74 & 57.78 & 87.51 & 78.01 \\ 
ReAct\cite{sun2021react}&\XSolidBrush    & 72.19 & 84.12&  53.17 & 88.12& 29.54 & 95.19 & 74.15 & 82.22 & 57.26 & 87.41 & 78.01 \\
    KNN \cite{sun2022knn}&\XSolidBrush& 51.01&89.46&41.12&90.55&7.32&98.50&54.08&88.31&38.38&91.71&78.01\\
    MaxLogit\cite{hendrycks2018deep}&\XSolidBrush&69.99&84.25&54.10&87.75&32.69&94.79& 71.42&82.79&57.05&87.40&78.01\\
    VOS\cite{du2022vos}&\Checkmark&43.03&91.92&56.67&87.64&31.65&94.53&41.62&90.23&43.24&90.86&\textbf{79.64}\\
    NPOS\cite{tao2023nonparametric}&\Checkmark&28.96&94.63&57.39&85.91&27.63&94.75&35.45&91.63&37.36&91.73&79.55\\
    AUTO&\XSolidBrush &\textbf{9.12} & \textbf{97.22} & \textbf{19.91} & \textbf{95.42} & \textbf{0.77} & \textbf{99.80} & \textbf{19.22} & \textbf{95.19} & \textbf{12.26} & \textbf{96.91} & 79.38\\
    \bottomrule
    \end{tabular}
    }
    \setlength{\belowcaptionskip}{-10pt}
   
    \label{imagenet}
\end{table*}

\textbf{AUTO performs more effectively while encountering larger ID space.}
Compared with methods that optimize models with auxiliary outliers, we notice that AUTO is not the best in evaluations on CIFAR-10 benchmarks.
However, as the complexity of the in-distribution (ID) space increases (from CIFAR-10 to CIFAR-100 and then to ImageNet-1k), the superiority of AUTO gradually becomes evident and establishes itself as the state-of-the-art method. In particular, the lead of AUTO over previous SOTA methods continues to expand (from 1.56\% (CIFAR-100) to 3.90\% (ImageNet-1k)).
The above phenomena suggest that the traditional OE paradigm can form compact decision boundaries in simple ID spaces to handle OOD data.
However, when the ID space becomes complex, previous paradigms struggle to maintain such compact decision boundaries, leading to significant OOD detection degradation.
On the contrary, AUTO adaptively adjusts decision boundaries specifically for the target out-of-distribution (OOD) data in deployment environments, leading to more efficient and better-performing OOD detection.

\textbf{AUTO effectively maintains the ID performance at test time.}
While significantly enhancing OOD detection performance, AUTO also mitigates the ID gradation effectively.
The largest gap in the three benchmarks is 1.48\% on ImageNet-1k with ResNet-50, which is acceptable.
AUTO does not impact the model's handling of the source ID task, which is practical for deploying real-world models.

\begin{figure}
    \centering
    \includegraphics[width=0.95\linewidth]{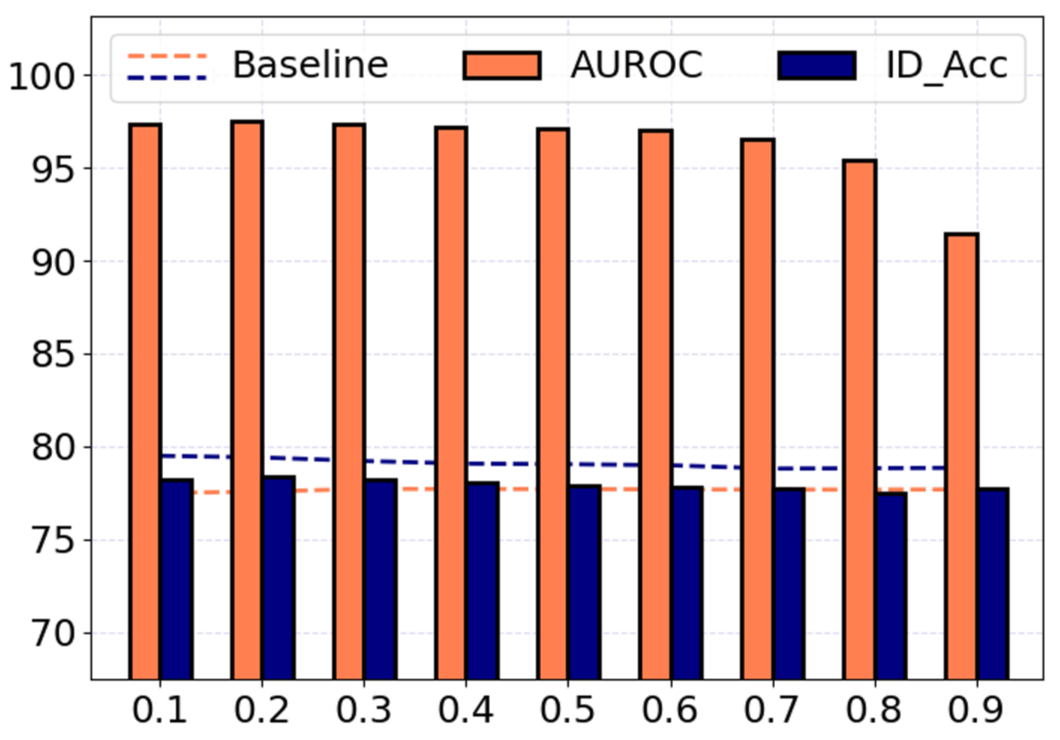}
   \caption{Effect of mixture ratio $\kappa_{t}$ on ID classification (orange) and OOD detection (blue).}
    \label{fig:dif-rat}
\end{figure}

\textbf{AUTO effectively handle OOD detection under varying ID to OOD data ratios.}
Going beyond the naive OOD detection setup, test-time OOD detection introduces a new setting where the mixed ratio $\kappa_{t}$ of ID to OOD data varies, testing the generality of the solution.
The performance of the AUTO method under different $\kappa_{t}$ conditions is shown in Figure~\ref{fig:dif-rat}. 
As a result, AUTO effectively maintains ID performances on different $\kappa_{t}$.
However, AUTO underperforms when there is very little OOD data.
We record the number of annotations for pseudo-OOD data and find that in scenarios with a higher proportion of OOD data, AUTO relies on continuous accurate annotation to achieve better OOD detection performance.
When there is less OOD data, accurate annotations significantly decrease, affecting the extent to which model performance is improved.

\textbf{AUTO achieves significant enhancement in complex-OOD scenarios.}
Except for various $\kappa_{t}$ scenarios, test-time OOD detection also involves complex OOD components at test time.
We perform experiments on multi-OOD scenarios and time-series scenarios, and the results are presented in Table \ref{tab:mix_data} and Figure \ref{fig:tso}, respectively.
In a word, models' performances in these new scenarios differ from an arithmetic average of performances in single-OOD scenarios.
The intricate composition of data presents challenges for all methods.
Nevertheless, AUTO continues to demonstrate exceptional performance, exhibiting a greater performance advantage over OE and WOODS.
This underscores AUTO's superior capability to handle mixed OOD scenarios.

\begin{table}[ht]
    \centering   
    \caption{Results on the mixed OOD scenarios. Models are ResNet-34 and ResNet-50, respectively.}
    \resizebox{\linewidth}{!}{
    
    \begin{tabular}{clrrr}
    \toprule
        \textbf{Data}& \textbf{Methods}&\textbf{FPR95 $\downarrow$} &\textbf{AUROC $\uparrow$}&\textbf{ID$\_$Acc $\uparrow$}\\
    \midrule
        CIFAR-100&MSP~\cite{hendrycks2016baseline}&82.69&75.07&77.51\\
        + &OE~\cite{hendrycks2018deep}&74.35&80.19&76.37\\
      Places365  &WOODS~\cite{katzsamuels2022training}&73.32&80.35&77.10\\
       SVHN &AUTO&\textbf{36.27}&\textbf{90.11}&\textbf{77.73}\\
    \midrule
    ImageNet&Energy~\cite{liu2020energy}&61.85&85.43&\textbf{76.12}\\
    +&DICE~\cite{sun2022dice}&32.42&92.22&\textbf{76.12}\\
    Places&DOE~\cite{wang2023outofdistribution}&63.34&81.98&75.02\\
    SUN&AUTO&\textbf{13.72}&\textbf{96.62}&73.08\\
    \bottomrule
    \end{tabular}
    }
    \setlength{\belowcaptionskip}{-12pt}
    \label{tab:mix_data}
\end{table}

\setlength{\tabcolsep}{1.0pt}
\begin{table*}[ht]
    \centering
    \caption{Comparison with competitive OOD detection methods in the NLP benchmarks.}
    \resizebox{\textwidth}{!}{

    \begin{tabular}{lcrrrrrrrrrrrrr}
    \toprule
    \multicolumn{1}{l}{\multirow{3}{*}{\textbf{Methods}}} &\multicolumn{1}{c}{\multirow{3}{*}{\textbf{$\mathcal{D}^{aux}$}}} &  \multicolumn{10}{c}{\textbf{OOD Datasets}} & \multicolumn{2}{c}{\multirow{2}{*}{\textbf{Average}}}&\multirow{3}{*}{\textbf{ID\_Acc}$\uparrow$}\\ 
    \cmidrule(lr){3-12}
    \multicolumn{2}{c}{ }   & \multicolumn{2}{c}{\textbf{SNLI}}   & \multicolumn{2}{c}{\textbf{IMDB}}  & \multicolumn{2}{c}{\textbf{Multi30K}} & \multicolumn{2}{c}{\textbf{WMT16}} & \multicolumn{2}{c}{\textbf{Yelp}} \\ 
    \cmidrule(lr){3-4}\cmidrule(lr){5-6}\cmidrule(lr){7-8}\cmidrule(lr){9-10}\cmidrule(lr){11-12}\cmidrule(lr){13-14}
    \multicolumn{2}{c}{}&  \multicolumn{1}{c}{FPR95$\downarrow$} &AUROC$\uparrow$  & FPR95$\downarrow$ &AUROC$\uparrow$  & FPR95$\downarrow$ &AUROC$\uparrow$  & FPR95$\downarrow$ & \multicolumn{1}{c}{AUROC$\uparrow$}  & \multicolumn{1}{c}{FPR95$\downarrow$ }&AUROC$\uparrow$ & \multicolumn{1}{c}{FPR95$\downarrow$ }&AUROC$\uparrow$ \\ 
    \midrule
    \multicolumn{15}{c}{\textbf{ID Data: 20 Newsgroups}}\\
   
    MSP\cite{hendrycks2016baseline}&\XSolidBrush& 37.75&86.32&62.66&80.47&58.79&78.43&45.43&85.40&65.65&78.72&54.06&81.87&73.25\\
    AUTO&\XSolidBrush&36.37&86.61&58.34&81.80&45.93&82.37&33.08&89.64&32.69&89.46&41.28&85.98&\textbf{73.44}\\
    OE (Guten)\cite{hendrycks2018deep}&\Checkmark&4.21&98.22&11.60&96.05&3.92&98.23&2.86&98.79&14.00&94.93&7.32&97.24&72.83\\
    OE (Wiki-103)\cite{hendrycks2018deep}&\Checkmark&2.29&98.85&2.91&98.66&2.90&99.66&0.34&99.73&76.51&75.45&16.59&94.47&72.34\\
    OE (Wiki-2)\cite{hendrycks2018deep}&\Checkmark&3.42&98.57&4.73&98.32&1.39&99.53&0.50&99.76&83.06&74.38&18.62&94.11&72.42\\
    OE+AUTO&\Checkmark&\textbf{2.36}&\textbf{98.89}&\textbf{2.91}&\textbf{98.70}&\textbf{0.86}&\textbf{99.65}&\textbf{0.33}&\textbf{99.76}&\textbf{3.43}&\textbf{98.48}&\textbf{1.98}&\textbf{99.10}&72.21\\
    \midrule
    \multicolumn{15}{c}{\textbf{ID Data: TREC}}\\
    
    MSP\cite{hendrycks2016baseline}&\XSolidBrush& 25.39&93.13&77.15&75.86&66.80&81.75&50.59&84.11&76.95&74.20&59.38&81.81&76.80\\
    AUTO&\XSolidBrush&17.58&94.96&47.85&88.29&2.27&93.01&42.58&87.31&24.02&93.07&30.86&91.33&\textbf{76.97}\\
    OE (Wiki-103)\cite{hendrycks2018deep}&\Checkmark& 19.73&93.20&0.98&99.45&3.32&99.75&10.35&96.79&0.39&99.89&6.95&97.69&63.20\\
    OE (Wiki-2)\cite{hendrycks2018deep}&\Checkmark&8.20&97.21&2.54&99.17&0.39&99.75&0.9 &99.52&\textbf{0.00}&\textbf{100.00}&2.42&99.07&70.60\\
    OE+AUTO& \Checkmark& \textbf{2.58}&\textbf{98.84}&\textbf{0.97}&\textbf{99.46}&\textbf{0.32}&\textbf{99.76} &\textbf{0.87}&\textbf{99.53}&\textbf{0.00}&\textbf{100.00}&\textbf{0.97}&\textbf{99.52}&76.72\\
\bottomrule
    \end{tabular}
    }
    \setlength{\belowcaptionskip}{-10pt}   
    \label{nlp_task}
\end{table*}

\textbf{AUTO possesses outstanding generality and compatibility.}
After assessing AUTO's performance on CV benchmarks, we conducted additional tests on NLP benchmarks, and the results are presented in Table \ref{nlp_task}.
AUTO consistently demonstrates outstanding performance in NLP evaluations, showcasing robust generalization capabilities across diverse modalities.
Furthermore, AUTO improves the performance of models, encompassing not only those trained in an ID manner but also those trained in an OE manner.
Extensive results suggest that AUTO does not conflict with previous out-of-distribution (OOD) detection methods; instead, it serves as a complementary strategy to enhance OOD detection performance when testing models strengthened during training.

\subsection{Component Analysis}
\label{abla}

\textbf{Impact of the core learning objectives.}
We evaluate the impact of different objectives, as presented in Table~\ref{obj_com}.
Our results demonstrate that training the model solely with an ID memory bank leads to similar performance as the method without optimization, indicating that optimizing on ID data alone does not effectively enhance OOD detection. Furthermore, while training on outliers alone improves OOD detection, it results in catastrophic forgetting, as evidenced by the decline in ID classification accuracy.
With the help of the ID memory bank, the model jointly updated by both ID and OOD samples already exhibits progress in both OOD detection and ID classification.
Besides, our prediction-aligning objective enhances both ID and OOD performance further.
\begin{table}[ht]
    \centering
    \caption{Ablation study on different combinations of objectives. Model is trained on CIFAR-100 with ResNet-34.}
    \begin{tabular}{cccrrr}
    \toprule
    $\mathcal{L}^{\textrm{id}}$&$\mathcal{L}^{\textrm{ood}}$&
    \multicolumn{1}{c}{$\mathcal{L}^{\textrm{PA}}$} &\textbf{FPR95 $\downarrow$} &\textbf{AUROC $\uparrow$}&\textbf{ID\_Acc $\uparrow$}  \\
    \midrule
    $\checkmark$&~&~&79.89&76.34&77.50 \\
    &$\checkmark$&&60.26&79.36&65.92\\
    $\checkmark$&$\checkmark$&&12.45&97.37&77.56\\
    $\checkmark$&$\checkmark$&$\checkmark$&\textbf{11.06}&\textbf{97.50}&\textbf{77.91}\\
    \bottomrule
    \end{tabular}    
    \label{obj_com}
\end{table}

\textbf{Impact of different optimization parameters.}
Results presented in Table~\ref{tab:opti} evaluate the efficacy of various optimization parameters.
On the one hand, taking both ID and OOD tasks into account, the performance of optimizing the last parameter block is superior.
On the other hand, models in open-world scenarios, particularly those engaged in online stream applications, need to notice the optimization efficiency.
We note that the inference time per sample is approximately 5ms.
AUTO necessitates only a modest 3.2x increase in processing time, which is tolerable.
Thus, we conclude that the utilization of the last parameter block as the optimization objective is a more efficient strategy.

\begin{table}[ht]
    \centering
    \caption{Ablation study on different modulation parameters. Model is trained on CIFAR-100 with ResNet-34.}
    \begin{tabular}{lrrrr}
        \toprule
        \multicolumn{1}{c}{\textbf{Modu. Para.}}& \textbf{FPR95} $\downarrow$&\textbf{AUROC} $\uparrow$&\textbf{ID\_Acc} $\uparrow$&\textbf{Time} \\
        \midrule
        No Para.& 83.53&74.34&77.51& 1x \\
        Block 1 & 77.50&78.36&77.64&1.8x\\
        Block 2 & 40.10&88.70&73.32&2.3x\\
        Block 3 & 17.72&95.48&72.40&2.9x\\
        Block 4 & \textbf{11.06}&\textbf{97.50}&\textbf{77.91}&3.2x\\
        BN      & 14.33&96.58&77.16&3.1x\\
        FC      & 77.92&78.76&78.59&1.5x\\
        All Para.& 13.26&97.07&76.95& 11.2x\\
        \bottomrule
    \end{tabular}
    \setlength{\belowcaptionskip}{-10pt}
    
    \label{tab:opti}
\end{table}

\textbf{Impact of different OOD scoring functions.}
We evaluate AUTO on different scoring functions, and results are shown in Table \ref{tab:ood scores}.
AUTO performs well in the logits space, energy space, and softmax space, with minimal differences in OOD and ID performance across the three spaces. This empirical observation indicates the feasibility and generality of AUTO and the test-time OOD detection paradigm.

\begin{table}[ht]
    \centering
    \caption{Ablation study on different OOD scoring functions. Model is trained on CIFAR-100 with ResNet-34.}
    \begin{tabular}{lrrr}
    \toprule
    OOD score &\textbf{FPR95 $\downarrow$} &\textbf{AUROC $\uparrow$}&\textbf{ID\_Acc $\uparrow$}  \\
    \midrule
    MSP~\cite{hendrycks2016baseline}&11.40&97.41&77.69 \\
    Energy~\cite{liu2020energy}&11.58&97.49&77.76\\
    MaxLogit~\cite{hendrycks2022scaling}&\textbf{11.06}&\textbf{97.50}&\textbf{77.91}\\
    \bottomrule
    \end{tabular}
    %
    
    \label{tab:ood scores}
\end{table}

\begin{figure*}[t]
    \flushleft
    \setlength{\abovecaptionskip}{10pt}
    \setlength{\belowcaptionskip}{-5pt}
    \subfloat[FPR95 on different $\lambda^{out}$]{
        \label{fig:subfig:1a}
        \includegraphics[width=0.245\textwidth]{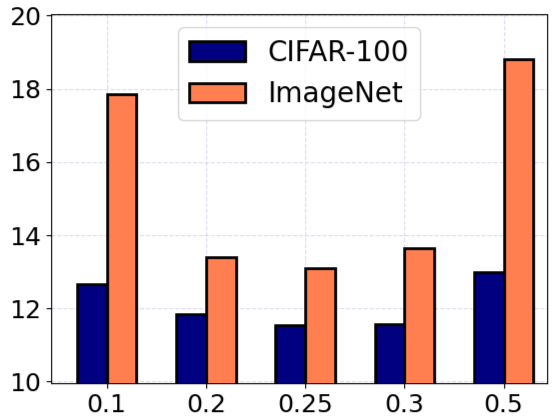}}
    \subfloat[FPR95 on different $k^{out}$]{
        \label{fig:subfig:1d}
        \includegraphics[width=0.245\textwidth]{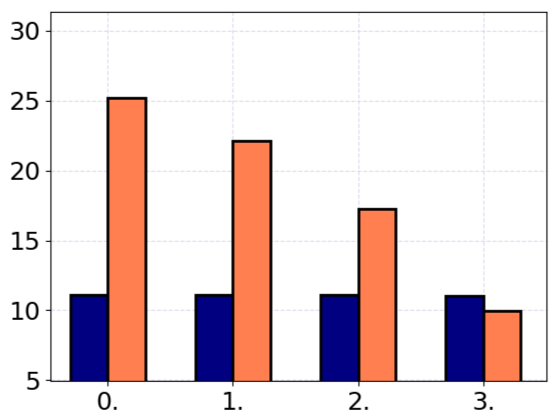}}
    \subfloat[FPR95 on different $\lambda^{PA}$]{
        \label{fig:subfig:1b}
        \includegraphics[width=0.245\textwidth]{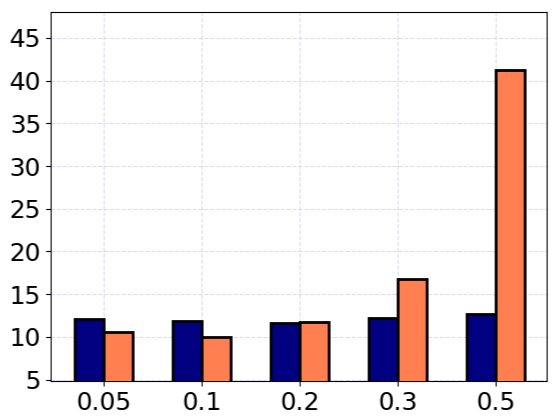}}
    \subfloat[FPR95 on different $\phi$]{
        \label{fig:subfig:1c}
        \includegraphics[width=0.245\textwidth]{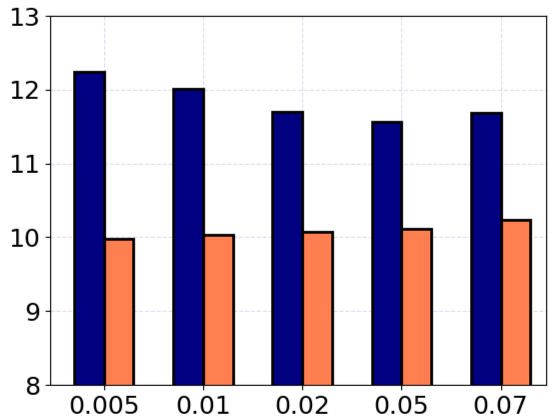}}
    \\
    \subfloat[ID\_Acc on different $\lambda^{out}$]{
        \label{fig:subfig:2a}
        \includegraphics[width=0.245\textwidth]{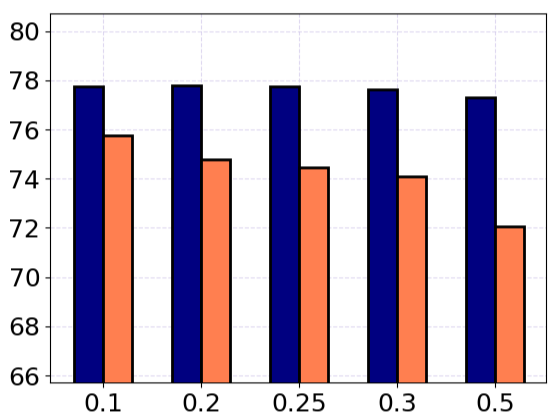}}
    \subfloat[ID\_Acc on different $k^{out}$]{
        \label{fig:subfig:2d}
        \includegraphics[width=0.245\textwidth]{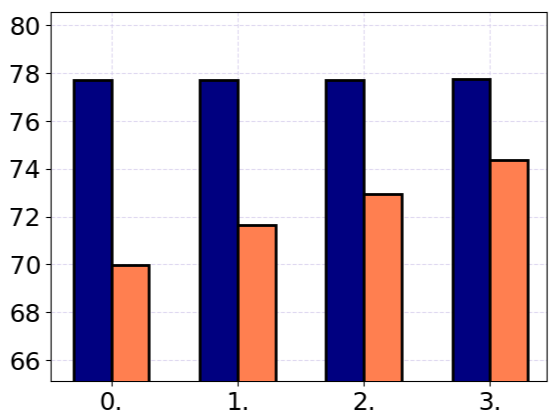}}
    \subfloat[ID\_Acc on different $\lambda^{PA}$]{
        \label{fig:subfig:2b}
        \includegraphics[width=0.245\textwidth]{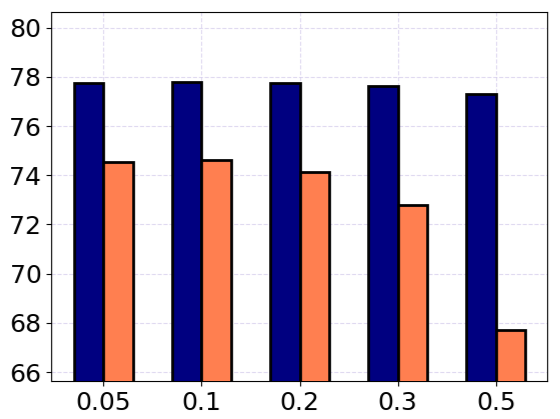}}
    \subfloat[ID\_Acc on different $\phi$]{
        \label{fig:subfig:2c}
        \includegraphics[width=0.245\textwidth]{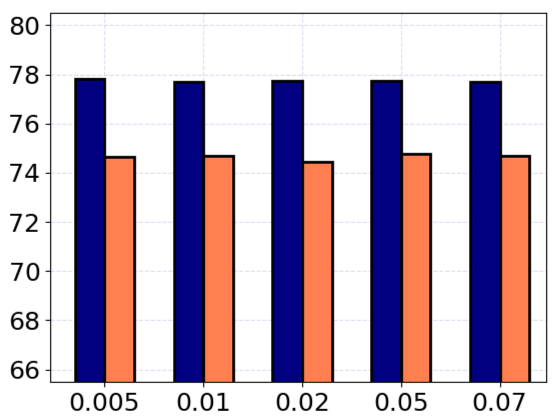}}   
    \caption{Performance of AUTO with varying $ \{ \lambda^{out}, \lambda^{PA}, \phi, k^{out} \}$ on ResNet. Average FPR95 and ID\_Acc are reported.}
    \label{fig:anti_exp}
\end{figure*}

\textbf{Impact of different memory design details.}
We design a class-wise dynamic ID memory bank in AUTO, which plays an important role in maintaining ID performance.
Results in Table \ref{tab:mem_type} show why we choose class-wise instead of random sampling and why we choose dynamic updating instead of fixed strategy.
(\textbf{Random Sampling}: The initialization of memory is done through random sampling, and with each memory update, the sample with the longest storage time is replaced.)
Considering the randomness of OOD sample occurrence, the number of times each sample is trained in the ID memory may vary. 
Random sampling of the memory can lead to different learning frequencies for samples from different ID categories. 
This, in turn, may cause the model to forget some ID semantics, leading to sub-optimal ID and OOD performances.

\begin{table}[h]
    \centering
    \caption{Ablation study on different ID memory details. Model is trained on CIFAR-100 with ResNet-34.}
    \begin{tabular}{lrrr}
    \toprule
    Memory &\textbf{FPR95 $\downarrow$} &\textbf{AUROC $\uparrow$}&\textbf{ID\_Acc $\uparrow$}  \\
    \midrule
    Random Sampling&15.24&96.38&77.53 \\
    Class-wise sampling&\textbf{11.06}&\textbf{97.50}&\textbf{77.91}\\
    \midrule
    Fixed &12.44&97.37&77.57\\
    Dynamic &\textbf{11.06}&\textbf{97.50}&\textbf{77.91}\\
    \bottomrule
    \end{tabular}
    \setlength{\belowcaptionskip}{-15pt}
    
    \label{tab:mem_type}
\end{table}


\subsection{Hyper-parameter Analysis}

\textbf{Impact of $\lambda^{out}$.} 
As shown in Figure \ref{fig:subfig:1a} and \ref{fig:subfig:2a}, an appropriate $\lambda^{out}$ is crucial for the proper functioning of AUTO. 
We empirically conduct that OOD regularization is insufficient when $\lambda^{out}$ is too small, and the OOD detection performance cannot be maximally improved.
In contrast, when $\lambda^{out}$ is too large,  the gradient changes significantly during one iteration, leading to rapid forgetting of the original ID knowledge.
Such forgetting leads to a mixture of ID and OOD data in the feature space, causing horrible ID and OOD performances.
In a word, $\lambda^{out}=0.25$ is the best option in our evaluations.

\textbf{Impact of $k^{out}$.}
The boundary for pseudo-OOD annotations is initialized by $k^{out}$, and Figure \ref{fig:subfig:1d} and \ref{fig:subfig:2d} show the impact of $k^{out}$ on OOD detection performance.
We recorded $\mu^{in}$ and $\sigma^{in}$ for ResNet-34, WRN-40-2 on CIFAR and ResNet-50, ViT-B-16 on ImageNet, respectively.
Results are shown in Table~\ref{tab:muandsigma}.
We observe that when $k^{out}$ is small, the pseudo-OOD annotation boundary is initialized to a mixed interval of OOD and ID, leading to sub-optimal performance as the model initially selects ID data as pseudo-OOD samples.
When $k^{out}$ is large, the number of accurately annotated OOD samples decreases significantly, reducing the iteration count, and resulting in sub-optimal performance due to underfitting.

\begin{table}[h]
    \centering
    \caption{ID Statistics for backbones.}
    \begin{tabular}{lcccc}
    \toprule
         Model& ResNet-34 &WRN-40-2  &ResNet-50 &ViT-B-16 \\
         \midrule
         $\mu^{in}$&0.9977&0.9554&0.8460&0.8034\\
         $\sigma^{in}$&0.0111&0.1248&0.2170&0.2435\\
         \bottomrule
    \end{tabular}
    
    \label{tab:muandsigma}
\end{table}

\textbf{Impact of $\lambda^{PA}$ and $\phi$.}
The extent of the prediction-aligning objective is controlled by the parameters $\lambda^{PA}$ and $\phi$, as shown in Figure \ref{fig:subfig:1b}, \ref{fig:subfig:2b} and Figure \ref{fig:subfig:1c}, \ref{fig:subfig:2c}.
$\lambda^{PA}$ is set to constrain the proportion of $\mathcal{L}^{PA}_t$ in the overall objective.
$\phi$ is set to constrain the extent of calibrations in test model predictions.
When $\lambda^{PA}$ or $\phi$ is small, the prediction-aligning regularization is insufficient to correct the model effectively.
In contrast, when $\lambda^{PA}$ is large, the prediction-aligning regularization overly focuses on anti-forgetting issues, weakening the effectiveness of the OOD regularization.
Meanwhile, when $\phi$ is large, the gradients of predictions are significantly changed, noticeably weakening the model's optimization towards a uniform vector in predictions for OOD data.
Considering that a large $\lambda^{PA}$ for $\mathcal{L}^{PA}_t$ can lead to underperformance on both ID and OOD tasks.
To address this issue, we propose a gradually decreasing weighting factor $\beta$, which decreases as the number of iterations increases.
Table \ref{tab:reduce} demonstrates that this factor effectively prevents the degradation of OOD performance, but it also reduces the gain of $\mathcal{L}^{PA}_t$ on ID performance.
Thus, we provide the following recommendations: if you prioritize ID performance, consider using $\mathcal{L}^{PA}_t$ without $\beta$. If you prioritize OOD detection performance, opt for the $\mathcal{L}^{PA}_t$ with $\beta$.
\begin{table}[h]
    \centering
    \caption{A gradually reducing weighting factor for $\mathcal{L}^{PA}_t$ can enhance OOD detection but reduces the gain of $\mathcal{L}^{PA}_t$ on ID performance. Models are trained on CIFAR-100 with ResNet-34 and tested on six OOD datasets.}
    \begin{tabular}{lrrr}
    \toprule
       $\mathcal{L}^{PA}_t$  & \textbf{FPR95} $\downarrow$& \textbf{AUROC} $\uparrow$ &\textbf{ID $\_$Acc} $\uparrow$\\
    \midrule
        with $\beta$ & \textbf{9.92}&\textbf{97.52}&77.84\\
        w/o $\beta$& 10.06&97.50&\textbf{77.91}\\
    \bottomrule
    \end{tabular}
    
    \label{tab:reduce}
\end{table}

Except for the individual hyper-parameter analyses mentioned above, we also surprisingly observe that the optimal hyper-parameters selected for each model consistently lead to excellent performance across different OOD test environments. 
This suggests that our hyperparameter selection is test-agnostic, meaning that for a given model, a fixed set of hyperparameters can be chosen to handle various deployment scenarios effectively.

\section{Conclusion}
In this paper, we propose the evolving test-time OOD detection problem where OOD detector is modifying online at test time.
Different from previous OOD detection setup, the new paradigm
Test-time OOD detection considers more practical and challenging scenarios.
We further propose a simple yet effective framework, AUTO, which adaptively selects and predicts test samples while updating models with them.
Extensive results demonstrate that our approarch can significantly enhance OOD detection performance while maintaining ID performance at the same time.
In the furture work, we plan to improve the annotation strategy thus enhancing the accuracy of selecting OOD samples in the unlabeled test data stream.
We hope our work could serve as a springboard for future works, provide new insights for revisiting the model development in OOD detection, and draw more attention toward the testing phase.


%


\ifCLASSOPTIONcaptionsoff
  \newpage
\fi



%
{
\bibliographystyle{IEEEtran}
\bibliography{ref}
}

\end{document}